\PassOptionsToPackage{square,numbers,sort}{natbib}
\documentclass{article}

\usepackage[final]{nips_2017}
\usepackage{tikz}
\usepackage{graphicx}
\usepackage{float}
\usepackage{subcaption}
\usetikzlibrary{shapes,shadows,arrows,fit,backgrounds,positioning}

\usepackage[utf8]{inputenc} % allow utf-8 input
\usepackage[T1]{fontenc}    % use 8-bit T1 fonts
\usepackage{hyperref}       % hyperlinks
\usepackage{url}            % simple URL typesetting
\usepackage{booktabs}       % professional-quality tables
\usepackage{amsfonts}       % blackboard math symbols
\usepackage{nicefrac}       % compact symbols for 1/2, etc.
\usepackage{microtype}      % microtypography

\title{Multiple-Instance, Cascaded Classification for Keyword Spotting in Narrow-Band Audio}

\author{
  Ahmad Abdulkader\\
  Voicera\\
  \texttt{ahmada@voicera.ai} \\
  \And
  Kareem Nassar \\
  Voicera \\
  \texttt{kareemn@voicera.ai} \\
  \AND
  Mohamed Mahmoud \\
  Voicera \\
  \texttt{geish@voicera.ai} \\
  \And
  Daniel Galvez \\
  Voicera \\
  \texttt{dt.galvez@gmail.com} \\
  \And
  Chetan Patil \\
  Voicera \\
  \texttt{chetanp@voicera.ai} \\
}

\begin{document}

\maketitle

\begin{abstract}
	We propose using cascaded classifiers for a keyword spotting (KWS) task on narrow-band (NB), 8kHz audio acquired in non-IID environments --- a more challenging task than most state-of-the-art KWS systems face. We present a model that incorporates Deep Neural Networks (DNNs), cascading, multiple-feature representations, and multiple-instance learning. The cascaded classifiers handle the task's class imbalance and reduce power consumption on computationally-constrained devices via early termination. The KWS system achieves a false negative rate of 6\% at an hourly false positive rate of 0.75.
	
\end{abstract}

\section{Introduction and Problem Description}
% Meetings are the crucibles where ideas are formed, decisions are made, and plans are set. Yet, they usually disappoint attendees as they feel unproductive: too much is said, and little is captured --- meetings tend to have a low signal-to-noise ratio (in the informal sense of the phrase).

% At Voicera, we are building Eva: an Enterprise Voice AI. Eva intelligently facilitates meetings by interacting with other participants --- using speech recognition --- to take note of decisions, action items, meeting minutes, etc. We believe Eva makes meetings more productive by distilling what really matters: highlights that users asked Eva to mark.

At Voicera, we are building Eva --- Enterprise Voice Assistant --- to collaborate with meeting participants using voice~\cite{voicera2017}. We believe interactions with Eva should feel as natural as interactions with any other participant in the meeting, so we have designed Eva to recognize and respond to voice commands. Eva continuously listens to the conversation during a meeting and verbally acknowledges the wake word "Okay Eva." In order for this interaction to feel natural, Eva’s KWS and audible responses need to be real-time.

Eva continuously predicts whether or not the keywords of interest were uttered in a real-time audio stream. For fear that users find Eva vexing due to unsolicited interruptions, the false positive rate should be less than 1 per hour; on the other hand, users may abandon the service if Eva doesn't respond when addressed, so we strive to maximize recall while maintaining a low false positive rate.

In addition to challenges that other KWS systems face while working with real-time speech, Eva's KWS system --- in order to be ubiquitous --- needs to support speech signals carried over the public-switched telephone network, which typically uses G.711: an NB audio codec that operates at a low bit-rate and at a sample rate of 8kHz~\cite{gallardo2015human}. Both human and automatic speech recognition suffer significantly from loss of accuracy when listening to NB audio~\cite{gallardo2015human,voran1997,moller2017}. Moreover, Eva's KWS system needs to adapt to new microphones, environment settings, speakers, and noise profiles whose characteristics vary drastically --- making the input signal to the KWS system non-IID.

\section{Prior Art}
Voice-enabled AI assistants --- like Apple's Siri, Amazon's Alexa, and Google Assistant --- perform similar KWS tasks to enable users to interact with their devices. Google proposed a model that uses Convolutional Neural Networks for performing a KWS task to detect 14 different phrases~\cite{sainath2015cnn}; this architecture showed a 27-44\% relative improvement in false positive rate compared to its predecessor that used a Deep Neural Network (DNN)~\cite{chen2014dnn}, which in turn showed a 45\% relative improvement in confidence score compared to KWS using Hidden Markov Models (HMM)~\cite{rohlicek1989}. In \cite{apple2017}, Apple uses a DNN to predict scores for 20 sound classes every 10ms, combining these scores in an HMM-like graph to calculate a composite score indicating that "Hey Siri" was uttered. Similar to our approach, which we detail below, the authors used a cascade of two classifiers to conserve power. Given the control Apple, Amazon, and Google have over their hardware specifications, they can acquire wide-band audio, which significantly improves the accuracy of speech recognition systems.

\section{Approach}
Our approach to the KWS problem is shown in Figure \ref{fig:approach}. The KWS system is triggered every 10ms. The audio signal from the past 500ms is captured for processing, since we found that the median command duration is 500ms. In this paper, we often call these 500ms windows "examples". Each of our six classifiers --- three cascaded classifiers composed of two DNNs trained on different feature representations --- are fully connected DNNs with two 128-input hidden layers. 
% The third phase of the pipeline uses a series of classifiers to categorize the set of fixed-feature representations into one of two categories: Keyword or Non-Keyword. The cascaded classifiers are only passed audio windows that not filtered out by preceding cascaded classifiers.
% In parallel, the audio signal is routed to an online training system []. The audio window examples collected are efficiently crowd-source-labeled []. The resulting supervised data is used to continue the training of the classifiers. The online training system deals efficiently with the unexpected variation in the audio signal distributions resulting from the use of a very wide variety of microphones and acoustic environments in which the audio signal is acquired.
\tikzstyle{decision} = [diamond, draw]
\tikzstyle{line} = [draw, -stealth, thick]
\tikzstyle{cir}=[draw, circle, minimum height=5mm, text width=3em, text centered]
\tikzstyle{block} = [draw, rectangle,text width=8em, text centered, minimum height=15mm, minimum width=1cm, node distance=7em]
\tikzstyle{smallblock} = [draw, rectangle,text width=0.9cm, text centered, minimum height=1.1cm,  minimum width=1.1cm,node distance=6.5em]

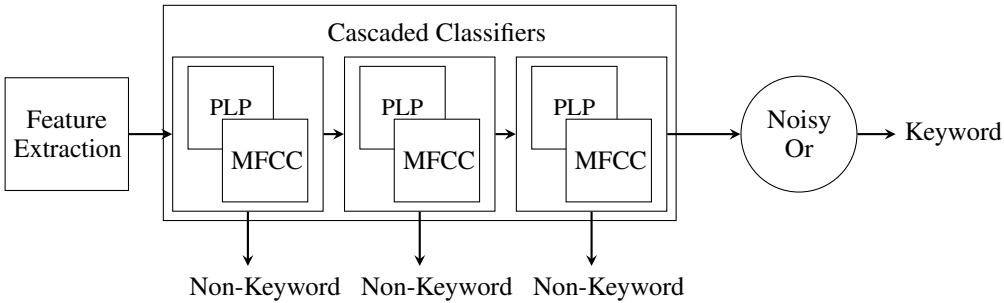
\begin{figure}[h]
   \setlength{\belowcaptionskip}{-10pt}
  \begin{tikzpicture}
  \node [block,text width=4em,xshift=-1.5em] (extraction) {Feature \\ Extraction};

  \node [smallblock,right of=extraction,xshift=1em,yshift=-1em,fill=white] (classifier1) {MFCC};
  \node [smallblock,right of=classifier1 ,fill=white] (classifier2) {MFCC};
  \node [smallblock,right of=classifier2 ,fill=white] (classifier3) {MFCC};
  
  \begin{scope}[on background layer]
     \node [smallblock,right of=extraction,xshift=-0.3em,yshift=1em,font=\small] (classifier_shadow) {PLP};
     \node [smallblock,right of=classifier1,xshift=-1.3em,yshift=2em,font=\small] (classifier_shadow2) {PLP};
     \node [smallblock,right of=classifier2,xshift=-1.3em,yshift=2em,font=\small] (classifier_shadow3) {PLP};
  \end{scope}

  \node [above of=classifier2,yshift=2em] (classifiers_label)  {Cascaded Classifiers};

  \begin{scope}[on background layer]
  
     \node[rectangle,draw,minimum width=2cm] [fit = (classifier1) (classifier_shadow),label=above:] (multi_rep1) {};
     \node[rectangle,draw,minimum width=2cm] [fit = (classifier2) (classifier_shadow2),label=above: ] (multi_rep2) {};
     \node[rectangle,draw,minimum width=2cm] [fit = (classifier3) (classifier_shadow3),label=above: ] (multi_rep3) {};
     \node[rectangle,draw,minimum width=2cm] [fit = (classifiers_label) (multi_rep1) (multi_rep2) (multi_rep3),label=above: ] (kws_system) {};
  \end{scope}
  
  \node [cir, right of=multi_rep3, xshift=5em] (noisy_or) {Noisy Or};
  
  \node [below of=multi_rep1,yshift=-2.5em] (non_keyword_area1)  { };
  \node [below of=multi_rep2,yshift=-2.5em] (non_keyword_area2)  { };
  \node [below of=multi_rep3,yshift=-2.5em] (non_keyword_area3)  { };

  \node [below of=classifier2,yshift=-2em] (non_keyword_area)  {Non-Keyword};
  \node [below of=classifier1,yshift=-2em] (non_keyword_area)  {Non-Keyword};
  \node [below of=classifier3,yshift=-2em] (non_keyword_area)  {Non-Keyword};
  \node [right of=noisy_or,xshift=3em] (keyword_area)  {Keyword};
  %\node[draw,ultra thick,text width=#2,align=left,#1]
  %\node [block,right of=extraction ] (classifier) {Cascade Classifiers};
  %\node [elli, above of=start, yshift=5em] (user) {User audio over telephony};
  %\node[block, below of=start, yshift=2em](decision1){Human labelling tool};
  %\node [block, left of=start, xshift=-5em] (utterance_detected) {Positive utterance detected};
  %\node [block, right of=decision1, xshift=5em] (retrain_nn) {Retrain NN};
  %arrows
  \path [line] (extraction) -- ([yshift=-2cm]multi_rep1);
  \path [line] (multi_rep1) -- (non_keyword_area1);
  \path [line] (multi_rep2) -- (non_keyword_area2);
  \path [line] (multi_rep3) -- (non_keyword_area3);
  
  \path [line] (multi_rep1) -- (multi_rep2);
  \path [line] (multi_rep2) -- (multi_rep3);
  \path [line] (multi_rep3) -- (noisy_or);
  \path [line] (noisy_or) -- (keyword_area);
  
%     \node [block,below of=extraction] (online_training) {Online training};
%   \path [line] (non_keyword_area) |- (online_training);
%   \path [line] (keyword_area) |- ([yshift=-2cm]online_training);
%   \path [line] (online_training) |- ([yshift=-1.75cm]kws_system);
  %\path [line] (user) -- (start);
  %\path [line] (start) -- node[anchor=south,text width=2cm,xshift=1em] {"Okay Eva" detected} (utterance_detected);
  %\path [line] (utterance_detected) |- node[anchor=south,xshift=5em] {playback audio feedback}(user);
  %\path [line] (utterance_detected) |- node[anchor=north,text width=3cm, xshift=5em] {chunk audio of utterance}(decision1);
  %\path [line] (decision1) -- node[anchor=north,text width=2cm] {TP and FP labeled}(retrain_nn);
  %\path [line] (retrain_nn) |- node[anchor=south,xshift=-5em] {deploy new model}(start);
  \end{tikzpicture}

  \centering
  \caption{KWS system diagram: (i) Feature Extraction (ii) Cascaded classifiers (iii) Noisy Or}
  \label{fig:approach}
\end{figure}

\subsection{Data}
We initially collected 19k user examples from 200 individuals using a crowdsourcing platform. The collection covered a wide variety of speakers living in the continental United States. These examples act as positive examples in our experiments. Negative examples for each cascaded classifier were generated from a repository of audio samples from a variety of meeting recordings that do not contain the keyword. All of the audio examples were either acquired as or converted to NB, 8kHz audio.

80\% of the data were designated for training; the remaining 20\% were used for evaluation. The data were stratified over individual speakers: A speaker belongs to either the training or the evaluation set.

\subsection{Multi-Representation Feature Extraction}
It is challenging to sample negative examples adequately. As such, discriminative classifiers suffer from the so-called "Novelty Detection" problem: a sound not encountered during the training process can be misclassified as a keyword --- resulting in a false positive. This happens because the decision boundary learned by the discriminative classifier is undefined in the areas that were not sampled during training.

To overcome this, we train two classifiers on two different representations, Mel Frequency Cepstral Coefficient (MFCC) and Perceptual Linear Prediction (PLP) features; both are commonly used in speech recognition~\cite{young2002htk}. These features are extracted on the same audio input for each stage of the cascade. These classifiers are ensembled using model-averaging to compute the final probability of an audio window being a keyword. Because these classifiers were trained on different representations of the same data, the classifiers concur on areas of the distribution that they were trained on and behave randomly otherwise. This results in a lower rate of false positives. In our system, both were extracted from 30ms frames with a stride of 10ms and concatenated for each 500ms example.

% Positive examples belonging to the keyword are reasonably easier to collect through crowd-sourcing data collection platforms. Even with the variations in accents and tones, generalization over all users can be overcome by collecting enough data.

% However, that is not the case for negative (non-Keyword) examples.

\subsection{Cascading}
Cascaded classifiers are commonly used to deal with highly asymmetric classification problems; they have been popularized through their use in many practical machine learning solutions; the Viola-Jones face detector is one notable example~\cite{viola2004robust}.

Cascading is an instance of ensemble learning based on the concatenation of several classifiers of increasing complexity. Each classifier is trained on the examples that are not filtered-out by the previous classifiers. A threshold is established on the output of each classifier such that a portion of the negative (non-keyword) examples it is subjected to get correctly classified. The remaining examples are either positive (keyword) examples or hard negative examples. For training the next classifier of the cascade, we run the previous classifiers of the cascade on a large repository of audio guaranteed not to have instances of the target keyword, using the discovered false positives as hard negative examples. This allows subsequent classifiers in the cascade to focus on negative examples that previous classifiers confuse with the positive examples. The distribution of positive to negative data gets more symmetric for subsequent classifiers. One drawback of this technique is that extracting hard examples for training subsequent classifiers in the cascade takes longer as the cascade gets better. In our experiments below, the first cascade classifier is trained with a ratio of 100 negative examples for every positive example. The second and third cascade classifier maintain a ratio of 2 negative examples for every positive example. During inference time, cascaded classifiers provide a way to early-terminate subsequent KWS computations on non-keyword windows.

\subsection{Multiple-Instance Learning}
The final stage of the pipeline, in Figure \ref{fig:approach}(iii), aggregates the outputs of the current and the past outputs of the cascaded classifiers to make a final decision about the triggering of the targeted keyword.
The KWS problem, as we modeled it, is innately a Multiple-Instance Learning (MIL) ~\cite{zhang2006multiple} problem: The keyword is somewhere in the audio signal, but at an imprecisely defined location and may vary drastically in duration as humans have a wide range of ways of pronouncing the same word. We've found that the "Okay Eva" utterance could vary from 300ms to 900ms.

In MIL, training examples are not singletons; they come in “bags” such that all of the examples in a bag share a label~\cite{zhang2006multiple,dietterich1997solving}. A positive bag of windows means that at least one window in the bag is positive while a negative bag means that all windows in the bag are negative. In MIL, learning must simultaneously learn which examples in the positive bags are positive, along with the parameters of the classifier. In our case, a bag is a group of 500ms windows strided by 10ms.

As such we have designed our learning process to learn simultaneously on all windows encompassed in a particular positive or negative bag of windows. The outputs corresponding to all of the windows in the bag are aggregated by a "Noisy Or."

\section{Experiments and Results}
\begin{figure}[H]
    \setlength{\belowcaptionskip}{-13pt}
    \centering
    \begin{minipage}[t]{0.48\linewidth}
        \centering
        \includegraphics[width=\linewidth]{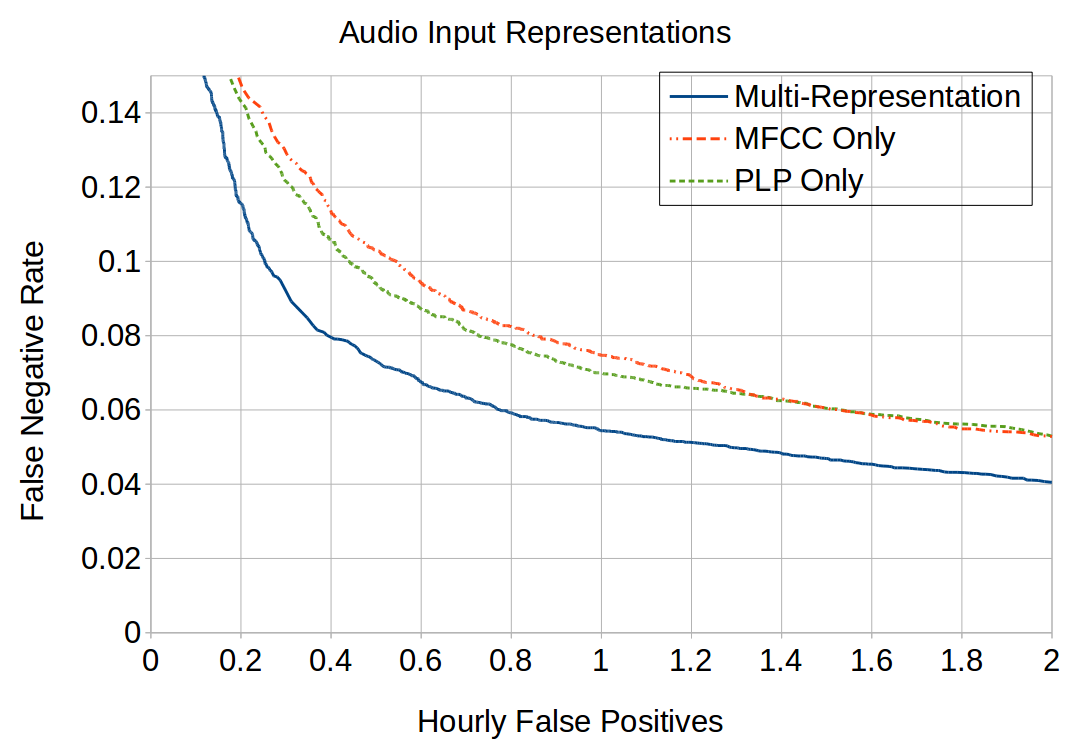}
        \small
        \caption{A plot showing the effects of using PLPs, MFCCs, and multi-representation models.}
        \label{fig:multi-rep}
    \end{minipage}
    \hfill
    \begin{minipage}[t]{0.48\linewidth}
        \centering
        \includegraphics[width=\linewidth]{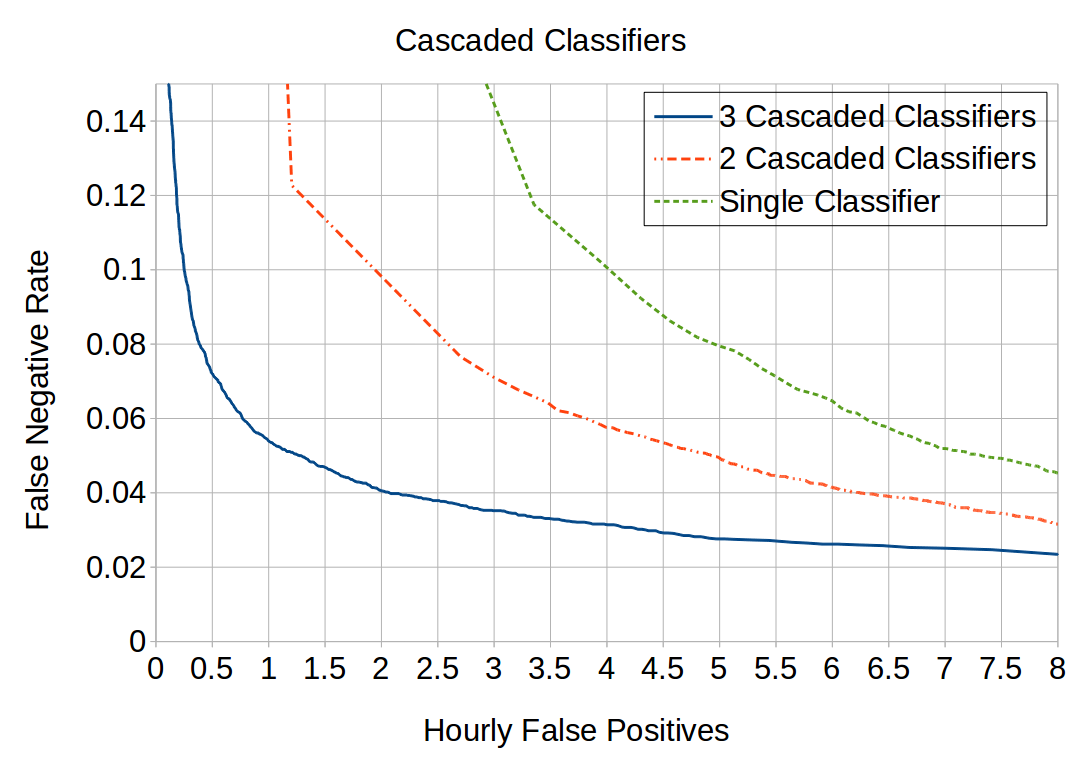}
        \caption{A plot showing the effects of using cascaded classifiers}
        \label{fig:cascaded}
    \end{minipage}
\end{figure}

We evaluated detection of the phrase "Okay Eva" using the hourly false positive rate plotted against the false negative rate.
% We have conducted our experiments on a number of keywords and voice commands. A "Wake Command": Okay-EVA, a "Sleep Command": Thanks-EVA and a number of other "Intention Commands": Action-Item for example. In the sections below, we will present the results for Okay-Eva. (EVA stands for Enterprise Voice Assistant [])
%number of metrics. The Precision/Recall graphs (a.k.a P/R or ROC), F1 measure, Area Under Curve (AUC) and Mean Time Between False Positives (MTBFP) were all used to assess the performance of Cascading, CNNs, Multiple Feature Representations and Multiple Instance Learning.
% \subsection{Voice Detection}
% Various RMS based heuristics were employed to perform voice-activity detection. These heuristics were tuned offline to filter-out as many non-voice regions as possible while allowing all voice regions. According to our statistics 72\% of the regions fall into the non-voice category. This helped improve the speed of the KWS significantly. 
\paragraph{Multiple-Feature Representations}
%MFCC and PLP features were extracted over 30ms windows, shifted by 10ms from each example to train two classifiers for each stage of the cascade. The output probabilities of of each of the classifiers are averaged. 
The ROC characteristics of a 3-stage cascade with only one feature representation (MFCCs or PLPs individually) compared to a 3-stage cascade with both representations is shown in Figure~\ref{fig:multi-rep}. The multi-representation scheme improves the trade-off between False Positives and False Reject rates. For example, at a False Reject rate of around 7\% (0.07), MFCCs have an hourly False Positive rate of 1.2; PLPs have an hourly False Positive rate of 1.0; the multi-representation has an hourly False Positive rate of 0.55.
\paragraph{Cascading}
%The initial set of classifiers in the first cascade were trained on a natural distribution of positive versus negative examples which resulted in a highly asymmetric distribution. 

We trained one-, two-, and three-cascaded classifiers for comparison. Because the first classifier is trained on mostly negative examples, the threshold on the output probability of the first classifier in each cascade was computed to guarantee that most of the positive examples would not be filtered out.
%For two- and three-stage classifiers, a ratio of two negative examples to one positive example was used for training. The distribution of these examples is not as asymmetric as the first training set. 
The ROC characteristics of each of the stages of the cascaded classifiers are shown in Figure~\ref{fig:cascaded}. The ROC characteristics significantly improves as more classifiers are cascaded.

%\begin{figure}[h!]
%  \centering
%  \begin{subfigure}[b]{0.4\linewidth}
%    \includegraphics[width=\linewidth]{roc.png}
%    \caption{Without cascading}
%  \end{subfigure}
%  \begin{subfigure}[b]{0.4\linewidth}
%    \includegraphics[width=\linewidth]{roc.png}
%    \caption{With cascading}
%  \end{subfigure}
%  \caption{Effects of cascading.}
%  \label{fig:coffee}
%\end{figure}

\section{Conclusion}

We demonstrate significant gains in KWS in narrow-band audio while minimizing computational resource usage. By incorporating multiple feature representations and three cascaded classifiers, we reduce our false positives per hour at 5\% false negative rate from 8 to 1.2, a reduction of 85\%. Although they are not directly comparable due to differences in datasets, our system performs better on narrow-band 8kHz audio than Google's DNN system~\cite{sainath2015cnn} performs on wide-band 16kHz audio.

%\section*{References}

%References follow the acknowledgments. Use unnumbered first-level
%heading for the references. Any choice of citation style is acceptable
%as long as you are consistent. It is permissible to reduce the font
%size to \verb+small+ (9 point) when listing the references. {\bf
%  Remember that you can go over 8 pages as long as the subsequent ones contain
%  \emph{only} cited references.}
\medskip

\small
\bibliographystyle{IEEEtranN}
\bibliography{references}
\end{document}